%% file: main.tex
\documentclass{svproc}
\usepackage{url}

\usepackage{import}
\usepackage{hyperref}
\usepackage{algorithm}
\usepackage{algorithmic}
\usepackage{textcomp}
\usepackage{float}
\usepackage{algorithm}
\usepackage{algorithmic}
\usepackage{mathtools}
\usepackage{commath}
\usepackage{fixmath}
\usepackage{import}
\usepackage{bbm}
\usepackage{times} 
\usepackage{amssymb}  
\usepackage{subcaption}
\usepackage{siunitx}

\usepackage{enumitem}
\setlist[itemize,1]{label=\textbullet}

\DeclareMathOperator*{\argmin}{\arg\!\min}

\begin{document}
\mainmatter              
\title{Momentum-Based Topology Estimation of \\ Articulated Objects}
%
%

\author{Yeshasvi Tirupachuri\inst{1}\inst{3} \and Silvio Traversaro\inst{2} \and  Francesco Nori\inst{4} \and Daniele Pucci\inst{2}}

%
%
%

\institute{RBCS, Istituto Italiano di Tecnologia, Genova, Italy
           \and
           Dynamic Interaction Control, Istituto Italiano di Tecnologia, Genova, Italy
           \and
           DIBRIS, University of Genova, Genova, Italy
           \and
           DeepMind, London, United Kingdom}

\maketitle              

\begin{abstract}
Articulated objects like doors, drawers, valves, and tools are pervasive in our everyday unstructured dynamic environments. Articulation models describe the joint nature between the different parts of an articulated object. As most of these objects are passive, a robot has to interact with them to infer all the articulation models to understand the object topology. We present a general algorithm to estimate the inherent articulation models by exploiting the momentum of the articulated system along with the interaction wrench while manipulating the object. We validate our approach with experiments in a simulation environment.



\end{abstract}

\import{sections/}{intro}
\import{sections/}{background}

\import{sections/}{method}

\import{sections/}{experiments}

\import{sections/}{results}

\import{sections/}{conclusions}
\import{sections/}{acknowledgements}
\import{sections/}{algorithm}

%
%

\bibliographystyle{unsrt}
\bibliography{main}

\end{document}

%% file: sections/intro.tex
\section{INTRODUCTION}
\label{introduction}

Over the last decade, there has been a growing interest in the robotics community to develop autonomous humanoid robots. Unlike laboratory settings, everyday environments are highly dynamic and unstructured. Articulated objects like doors, drawers, valves, and tools are multi-link rigid body systems with their object parts moving relative to one other. Articulation models describe the joint nature between two object parts. So, for a humanoid robot to operate autonomously in dynamic environments, it has to learn the articulation models. This paper contributes to learning articulation models and estimates the topology of articulated objects.

Doors are the most likely experienced articulated objects in many robotic applications like rescue scenarios, elderly care, hospitality, and others. The earliest investigations tackling the door opening problem are carried in \cite{nagatani1995experiment} and \cite{niemeyer1997simple}. The authors in \cite{nagatani1995experiment} assume a known door model and leveraged the combined motion of the manipulator and the autonomous mobile platform to open the door. In contrast, a model-free approach of controlled interactions along the path of least resistance is investigated in \cite{niemeyer1997simple}. Later, the concept of equilibrium point control (EPC) for the specific task of opening novel doors and drawers is evaluated in \cite{jain2009pulling}. In addition, they implemented an articulation model estimation algorithm using the end-effector trajectory, assuming a stable grasp and planar motion of the end-effector. The algorithm returns an estimate of the rotation axis location and the radius. The prismatic joint is estimated as a rotational joint with a large radius. More recently, a model-free adaptive velocity-force/torque controller for simultaneous compliant interaction and estimation of articulation models in objects like doors and drawers with one degree of freedom motion is proposed in ~\cite{karayiannidis2016adaptive}. Additionally, they provide proof of convergence of the articulation model estimates.

On the other hand, the idea of interactive perception paradigm is introduced in \cite{katz2007interactive} and \cite{katz2008manipulating} highlighting the need for extracting task-specific perceptual information using the manipulation capabilities of a robot by interacting with the environment. They employ optical flow based tracking of features on moving object parts and build a graph. Then the articulation models are extracted from the information contained in the graph. The rotational joint is identified by rotating centers between two sub-graphs and prismatic joint by shifting movements of sub-graphs. They successfully demonstrated the use of interactive perception in extracting the kinematic model of various tools to build a Denavit-Hartenberg (DH) parameter model and then use it to operate a tool. In addition, a symbolic learning-based approach to manipulation is presented in \cite{brock2009learning} which uses relational representations of kinematic structures that are grounded using perceptual and interaction capabilities of a robot. They successfully demonstrated learning and generalization of manipulation knowledge to previously unseen objects.

A probabilistic learning framework proposed in \cite{sturm2009learning} uses a noisy 3D pose observations of object parts. They implemented predefined candidate joint models with parameters and also a non-parametric Gaussian process model to which observed 3D pose trajectory data of object parts is fit to find kinematic structures of kinematic trees. Later, a stereo camera system is used to get dense depth images as input \cite{sturm20103d}. Building on the previous work, a unified framework with several extensions like dealing with kinematic loops and an extended set of experiments is presented in \cite{sturm2011probabilistic}. A particle filter based approach presented in \cite{hausman2015active} integrates the idea of interactive perception into a probabilistic framework using visual observations and manipulation feedback from the robot. They also presented best action selection methods based on entropy and information gain which guides the robot to perform the most useful interactions with the object to reduce the uncertainty on articulation model estimates.

The concept of exploration challenge for robots where the task is to perform explorative actions and learn the structure of the environment is presented in \cite{otte2014entropy}. One of their main contributions is probabilistic belief representation of articulation models including properties like friction and joint limits. They successfully demonstrated how the behavior emerged from entropy-based exploration is more informative than explorative strategies based on heuristics. An online multi-level recursive estimation algorithm considering task-specific priors based on the concept of interactive perception is presented in \cite{martin2014online}. They use a series of RGB-D image data as input to estimate articulation models including the joint configuration. Further, they extended their approach \cite{martin2017building} integrating information from vision, force-torque sensing and proprioception. In addition to kinematic articulation model estimation, they also generated a dynamic model of the articulated object. 

In this paper, we propose an algorithm to estimate the topology of a complex floating base articulated object by leveraging the momentum and interaction wrench information while manipulating the object. Unlike the previous approaches, our approach is addressed to handle floating base objects. Further, our method attempts to identify the topology of an articulated system with any number of degrees of freedom. This paper is organized as follows. Section~\ref{background} introduces the notation and the problem statement. Section~\ref{method} presents our method and algorithm. Section~\ref{experiments} provides the details of the experiments. Section~\ref{results} contain the numerical results showing the articulation model estimation followed by conclusions.

%% file: sections/background.tex
\section{BACKGROUND}
\label{background}

Spatial vectors \cite{featherstone2014rigid} are 6D vectors that are proven to be powerful tools in analyzing rigid-body dynamics. Unlike the standard notation of spatial vectors, we use a modified notation. In the case of spatial motion vectors, we consider the linear part first followed by the angular part and in the case of spatial force vectors, we consider the forces first followed by the moments.

\subsection{Notation}
\begin{itemize}
  \item $A$ denotes the inertial frame, $B$ denotes a body-fixed frame and $com$  denotes a frame associated with the center of mass of a rigid body.
  \item Let $u$ and $v$ be two~$n$-dimensional column vectors of real numbers, i.e. $u$, $v$ $\in$ $\mathbb{R}^n$, their inner product is denoted as $u^T$$v$, with $T$, the transpose operator.
   \item $SO(3)$ denotes the set of \ $\mathbb{R}^{3\times3}$ orthogonal matrices with determinant equal to one.
   \begin{align*}
    SO(3):= \{\, R \in \mathbb{R}^{3 \times 3} \mid R^T R = I_3 , \hspace{0.3em} \operatorname{det}(R) = 1 \,\}
   \end{align*}
   \item Given $u$, $v$ \ $\in$ \ $\mathbb{R}^3$, \ $S(u)$ \ $\in$ \ $\mathbb{R}^{3\times3}$ denotes the \textit{skew-symmetric} matrix-valued operator associated with the cross product in \ $\mathbb{R}^{3}$, \ such that \ $S(u)v \ = u \times v$.
   \item Given the vector $u = (x;y;z) \in \mathbb{R}^3$, we define the skew-symmetric matrix as,
   \begin{align*}
     S(u) = \begin{bmatrix}
              0 && -z && y \\
              z && 0 && -x \\
              -y && x && 0
            \end{bmatrix}
   \end{align*}
   \item $||u||$ denotes the \textit{euclidean norm } of a vector, \ $u \ \in \mathbb{R}^3$.
   \item $\mathbf{S} \in \mathbb{R}^{{n_{f}} \times 6}$ is the motion subspace matrix \cite{featherstone2014rigidCh3} of a joint, that has $n_f$ degrees of freedom and $q \in \mathbb{R}^{6}$ is a column vector that denotes the joint variable.
   \item $1_n$ $\in$ $\mathbb{R}^{n \times n}$ denotes the identity matrix of dimension~$n$.
   
   \item $p_\textsc{B} \in \mathbb{R}^{3}$ denotes the origin of the frame $B$, expressed in the inertial frame; $^AR_B \in SO(3)$ is the rotation matrix that transforms 3D vector, expressed with the orientation of $B$ to a 3D vector expressed in frame $A$.
   \item $P \in \mathbb{R}^{7}$ denotes the 3D pose of a rigid body with respect to the inertial frame A
   \begin{align*}
     P = \begin{bmatrix}
           p_\textsc{B} \\
           \mathrm{q}
         \end{bmatrix}
   \end{align*}
   where $\mathrm{q} \in \mathbb{R}^{4}$ denotes the orientation of the rigid body expressed as a quaternion
   \item $\omega \in \mathbb{R}^{3}$ denotes the angular velocity of a rigid body,  expressed in the body frame $B$, defined as
   \begin{align*}
     S(\omega) = \ ^AR_B^T \ ^A\dot{R}_B
   \end{align*}
   \item $\mathrm{v}$ $\in \mathbb{R}^{6}$ denotes the twist of a rigid body, expressed in the body frame $B$,
   \begin{align*}
     \mathrm{v} = \begin{bmatrix}
           ^A{R}^{T}_{B} ~ \dot{p}_\textsc{B} \\
           \omega
         \end{bmatrix}       
   \end{align*}        
   \item $\mathrm{f} \in \mathbb{R}^{6}$ denotes an external wrench exerted on the body, expressed in the body frame $B$
   \item \text{g} $\in \mathbb{R}^{6}$ denotes the gravitational force vector
   \item $\mathrm{M} \in \mathbb{R}^{6 \times 6}$ denotes the spatial inertia, expressed in the body frame $B$
   \begin{align*}
     \mathrm{M} = \begin{bmatrix}
           m 1_3 && -m S(c)\\
           m S(c) && I_B 
         \end{bmatrix}
   \end{align*}
   \begin{itemize}
     \item $m \in \mathbb{R}$ denotes the mass of a rigid body,
     \item $c \in \mathbb{R}^{3 \times 3}$ denotes the center of mass of a rigid body, expressed in the body frame $B$
     \item $I_B \in \mathbb{R}^{3 \times 3}$ denotes the 3D rotational inertia matrix of a rigid body, expressed with the orientation of the body frame $B$ and with respect to the origin of the body frame $B$
     \item $I_c \in \mathbb{R}^{3 \times 3}$ denotes the 3D rotational inertia matrix of a rigid body $B$, with respect to the center of mass of the body, where:
     \begin{align*}
       I_B = I_c - mS(c)S(c)
     \end{align*}        
   \end{itemize}
   \item $h = \mathrm{M} \mathrm{v} $ denotes the spatial momentum of a rigid body with respect to the body frame $B$
   \item $X_B \in \mathbb{R}^{6 \times 6}$ denotes spatial transformation from frame B to the inertial frame $A$
   \item $H_B \in \mathbb{R}^{4 \times 4}$ denotes homogeneous transformation from frame B to the inertial frame $A$
   \item Operator $H(\cdot) : \mathbb{R}^7 \rightarrow \mathbb{R}^{4 \times 4}$ takes 3D pose and returns homogeneous transformation matrix.
   \item Operator $X(\cdot)  : \mathbb{R}^{4 \times 4} \rightarrow  \mathbb{R}^{6 \times 6}$ takes a homogeneous transform as input and returns a spatial transformation.
   \item Operator $lin(\cdot) : \mathbb{R}^{4 \times 4} \rightarrow \mathbb{R}^3$ takes a homogeneous transformation matrix as input and returns the position.
   \item Operator $rot(\cdot) : \mathbb{R}^{4 \times 4} \rightarrow \mathbb{R}^{3 \times 3}$ takes a homogeneous transformation matrix as input and returns the rotation matrix.
\end{itemize}

\subsection{Problem Statement}

\begin{figure}[h]
  \centering
  \includegraphics[scale=0.4]{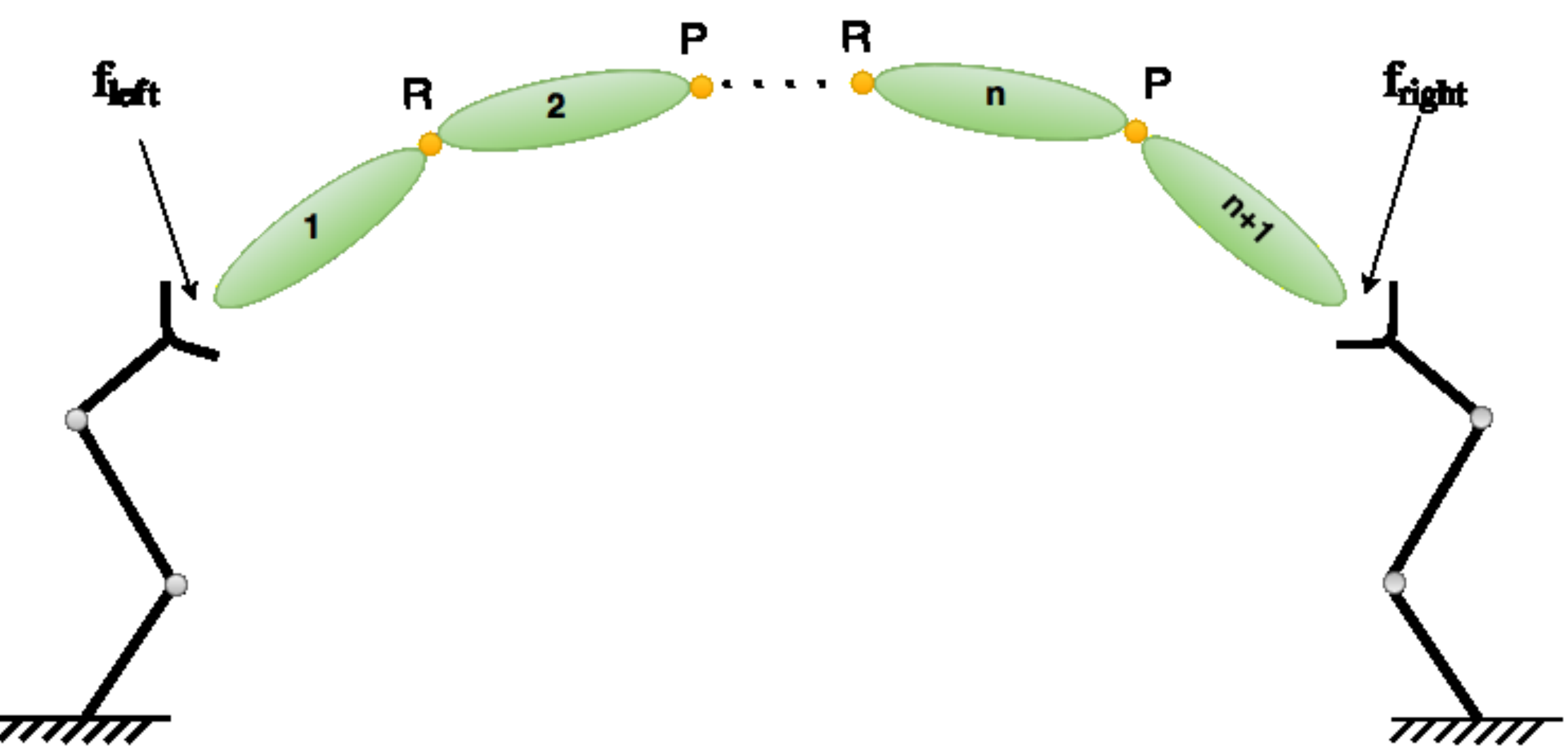}
  \caption{Articulated object manipulation}
  \label{articulated-object-manip-problem-statement}
\end{figure}

Consider a floating base articulated object as shown in Fig.~\ref{articulated-object-manip-problem-statement} with $n+1$ rigid bodies called links. The links are connected to one another by one degree of freedom articulation model. We assume to have the simplest articulation models of either a revolute joint model $(R)$ or a prismatic joint model $(P)$. We define the set of joint indices, $J = \{ 1,2,...,n \}$ and the set of articulation models, $\mathbb{M} = \{R,P\}$. Now, the topology of the articulated object is represented by the set $\Delta$, whose elements are pairs of elements from the sets $J$ and $\mathbb{M}$ i.e

\begin{equation}
    \Delta = \{ \{1,\mathbbm{m}_1 \}, \{2,\mathbbm{m}_2 \},....,\{n,\mathbbm{m}_n \} \}
    \label{eq:Delta}
\end{equation}

where, $\mathbbm{m}_1,\mathbbm{m}_2,....,\mathbbm{m}_n \in \mathbb{M}$

The articulated object is assumed to be of a serial chain kinematic structure. An anthropomorphic robot with two arms manipulate the object by holding the terminal links which result in the interaction wrenches $\mathrm{f}_{left}$ and $\mathrm{f}_{right}$ at the arms of the robot. The contacts between the terminal links and the arms of the robot are considered rigid. Now, the problem we are interested in is to leverage the kinematic evolution of the links and the interactions wrench to estimate the set $\Delta^*$ that represents the true articulation models present in the object

%% file: sections/method.tex
\section{METHOD}
\label{method}

Consider a complex articulated object as shown in Fig. \ref{articulated-object-manip-problem-statement}. The momentum of the $i$-th rigid body expressed in its body frame, is given by,

\begin{equation}
    \mathrm{h}_i = \mathrm{M}_i \mathrm{v}_i
\end{equation}

In a kinematic tree structure, the twist of the $i$-th rigid body, expressed in its body frame, is given by,
\begin{equation}
  \mathrm{v}_i = \  ^iX_{i-1} \ \mathrm{v}_{i-1} + \mathrm{v}_{J_{i-1}}
  \label{generalBodyVel}
\end{equation}
where,
  \begin{itemize}
    \item $^iX_{i-1}$ is the spatial transformation from the parent link to the child link.
    \item $\mathrm{v}_{J_{i-1}}$ is the twist of the $i-1$-th joint, connecting link $i$ to its parent, expressed in the body frame of link $i$.
  \end{itemize}

The twist of a joint, expressed in the child link body frame, is given by,
\begin{equation}
    \mathrm{v}_{J_{i-1}} = \mathrm{S_{i-1}} \ {\dot{q}}_{i-1}
    \label{jointVel}
\end{equation}

Now, the joint twist depends on the nature of the articulation \textit{model} present between the two links that are connected by the joint and can be written as,
\begin{equation}
    ^{model}{\mathrm{v}_{J_{i-1}}} = \ ^{model}\mathrm{S_{i-1}} ~^{model} \ {\dot{q}}_{i-1}
    \label{jointVelModel}
\end{equation}

Following the relations (\ref{jointVel}) and (\ref{jointVelModel}), we can express the momentum of the $i$-th rigid body in terms of the articulation \textit{model} present between it and its parent link. In this way, we encode the articulation model information in the momentum of a rigid body.

\begin{equation}
    ^{model}\mathrm{h}_i = \mathrm{M}_i ~^{model}\mathrm{v}_i
\end{equation}

The net wrench acting on any $i$-th rigid body expressed in the body frame is the gravitational wrench give by,

\begin{equation}
    W_i = m_i \ \text{g}
    \label{gravity-wrench}
\end{equation}

In addition, the terminal links experience reaction wrenches $-\mathrm{f}_{left}$ and $-\mathrm{f}_{right}$ from the arms of the robot. So, the total net wrench acting on the articulated object is given by,
\begin{equation}
    W = \ - {^AX}_{left}^* \ \mathrm{f}_{left} - {^AX}_{right}^* \ \mathrm{f}_{right} + \sum\limits_{i=1}^n \ {^AX}_{{com}_i}^* m_i \ \text{g}
    \label{generalTotalWrench}
\end{equation}
where $^AX^*$ is the spatial transformation for force vectors with respect to the inertial frame A. The total momentum of the articulated object is equal to the sum of its link momenta given by,

\begin{equation}
    ^{\Delta}h = \sum\limits_{i=1}^n \ ^A{X}^*_i \ ^{model}{\mathrm{h}}_i
\end{equation}

where $\Delta$ represents the topology of the articulated object.

According to classical mechanics \cite{featherstone2014rigidCh2}, the net wrench $W$, acting on a rigid body system is equal to the rate of change of its momentum expressed with respect to the inertial frame of reference, $A$.
\begin{equation}
    W = \ ^{\Delta}{\dot{h}}
    \label{wrench-hdot}
\end{equation}

Now, for $n$ number of joints, we will have $2^n$ sets. The set, $\Delta^*$ which solves the following optimization represents the true topology of the articulated object.
\begin{equation}
    {\Delta}^*= \argmin\limits_{\Delta_j}\sum\limits^{2^n}_{j=1} ||W - {^{\Delta_j}{\dot{h}}}||
\end{equation}

%% file: sections/experiments.tex
\section{EXPERIMENTS}
\label{experiments}

\begin{figure*}[h]
    \centering
    \begin{subfigure}[t]{0.5\textwidth}
        \centering
        \includegraphics[width=0.95\textwidth]{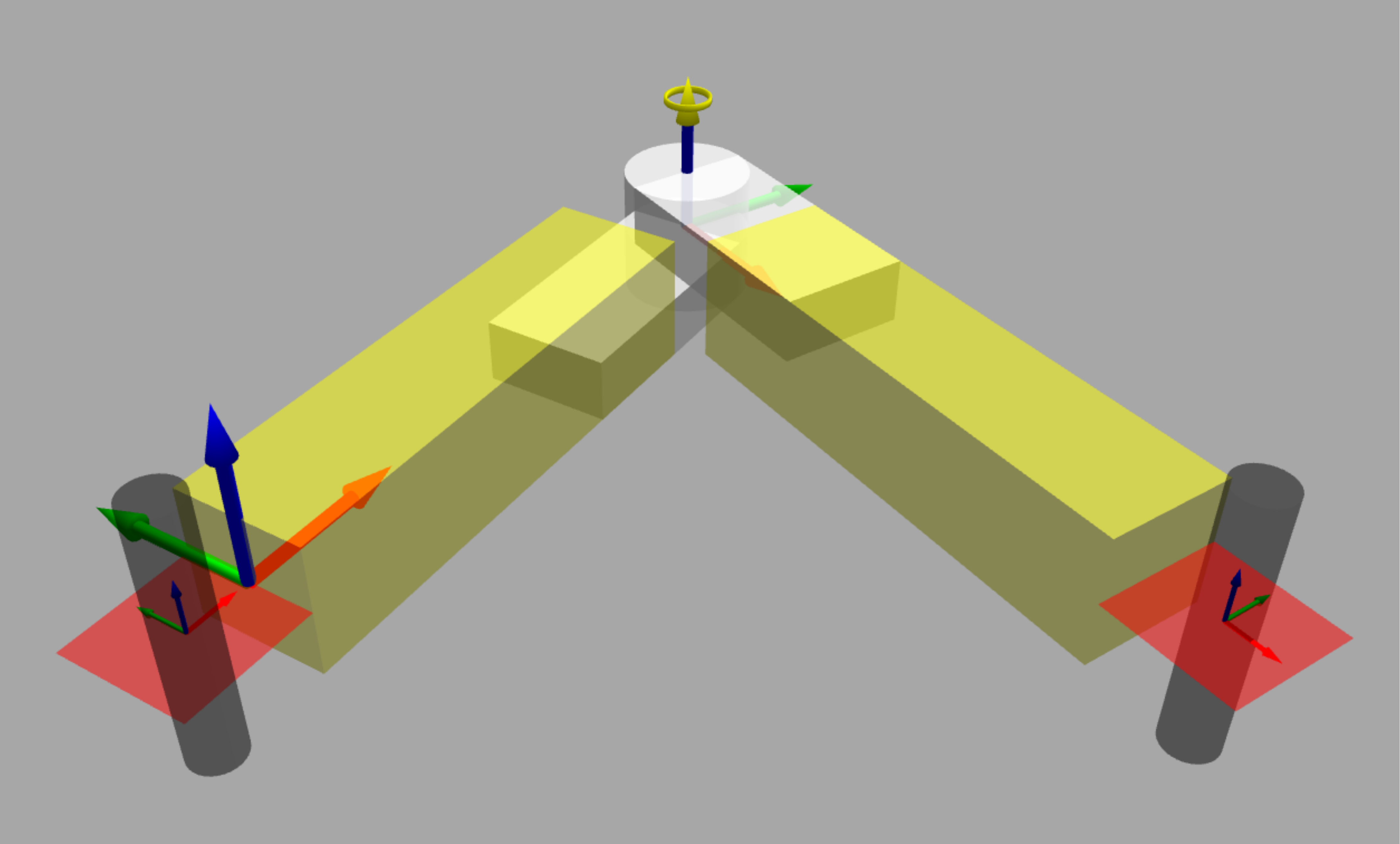}
        \caption{Revolute Model}
		\label{pmodel}
    \end{subfigure}%
    ~ 
    \begin{subfigure}[t]{0.5\textwidth}
        \centering
        \includegraphics[width=0.95\textwidth]{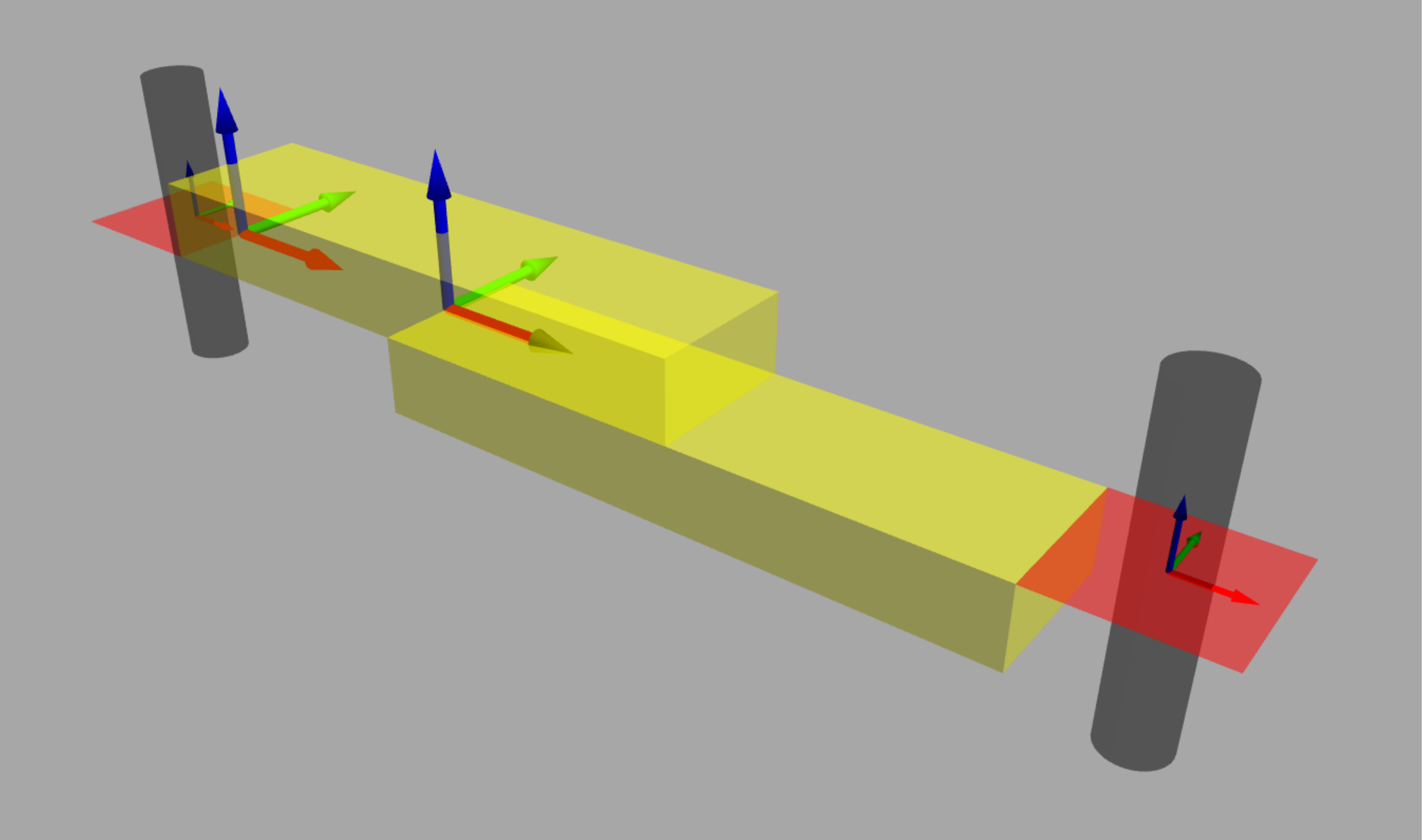}
        \caption{Prismatic Model}
		\label{rmodel}
    \end{subfigure}
    \caption{Articulated Objects Model}
    \label{fig:articulated-objects}
\end{figure*}

As a proof of concept experiment, we considered simple articulated objects as shown in Fig. \ref{fig:articulated-objects} containing two links connected through either a revolute joint (Fig. \ref{rmodel}) or a prismatic joint (Fig. \ref{pmodel}). The motivation behind this experimental choice is that many real-life articulated objects like scissors, pliers, drawers and other articulated objects can be represented in this simple form. Accordingly, we modeled two objects in gazebo simulation environment using  Simulation Description Format (SDF). The revolute model articulated object contains two links connected through a revolute joint and the prismatic model articulated object contains two links connected through a prismatic joint. The joints are designed with a damping value of $0.1$ and static friction value of $0.1$.

\begin{figure}[h]
  \centering
  \includegraphics[scale=0.25]{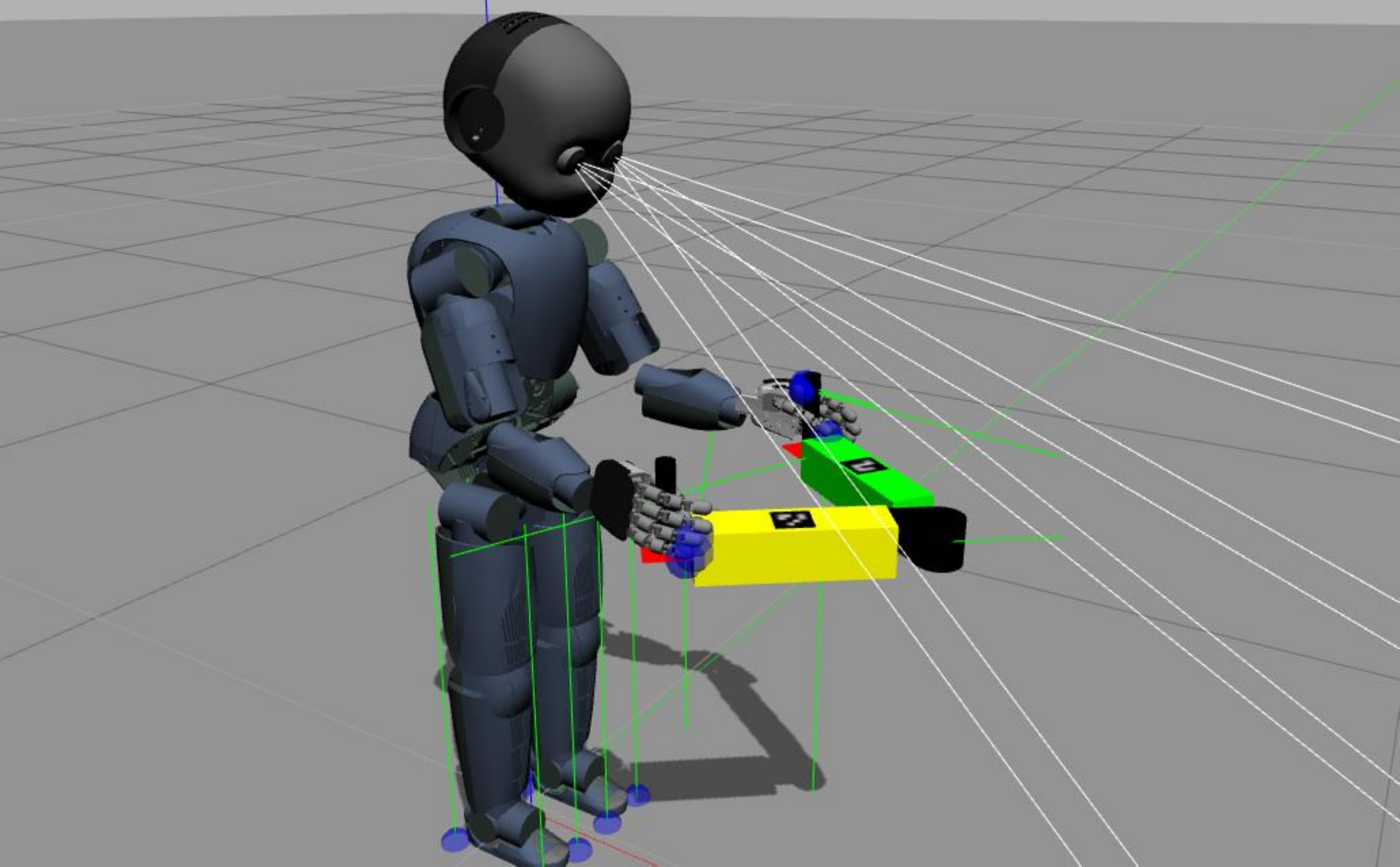}
  \caption{Scenario of iCub robot manipulating an articulated object }
  \label{icubmanip}
\end{figure}

We envision an experimental scenario where the humanoid robot iCub \cite{metta2010icub} \cite{Nataleeaaq1026} will hold the articulated object, as shown in Fig. \ref{icubmanip} and perform exploratory actions to estimate and learn the topology of the object.

The cylindrical elements in black color are the handles of the terminal links. They are designed to be virtual links without any significant mass and inertial values to contribute towards the system dynamics. The rectangular elements in yellow color are the object links that are connected to handles through fixed joints. 

Several real-world articulated objects are passive and do not contain any sensors to give the information related to the motion of the links or the wrenches acting at the terminal links. A vast amount of research has been carried on tracking rigid bodies either using markers, features or depth information, yet the problem of obtaining robust 3D pose values of rigid bodies is still an open challenge in the field of computer vision. As visual perception is not the main goal of this work, we acquire the pose values directly from the simulation environment using a plugin. Also, we made the assumption to have full knowledge of the link inertial parameters i.e., mass, inertia, and center of mass. In the case of iCub robot, external wrenches acting at the hands are estimated using the techniques developed for whole-body control \cite{nori2015icub}. So under the assumptions of rigid contacts between the terminal links of the articulated object and the arms of the robot, the wrenches acting on the terminal links of the object are simply the reaction forces from the arms of the robot. 

In this proof of concept, we primarily want to highlight the articulated motion estimation approach. Operating on an articulated object by a humanoid robot without its true object model poses quite a challenge on the control aspects of the experiment. So, we did not consider the iCub robot to manipulate the articulated object. Also, we embedded a simulated 6 axis Force-Torque sensor plugin \cite{hoffman2014yarp} at the handles to measure the external wrenches acting on the object terminal links. Furthermore, one of the handles is anchored to the world in gazebo simulation through a fixed joint and this also anchors the object link attached to that handle. The other link is free to move and we apply an external sinusoidal \textit{exploration wrench} of random frequency and amplitude mimicking the exploratory actions a robot performs while manipulating the object without being certain of the articulation models.

%% file: sections/results.tex
\section{RESULTS}
\label{results}

The range of motion for the prismatic joint is set to $\SI{0.15} \meter$  and for the revolute joint, $95^\circ$. The amplitude range of the exploration wrench is $[\SI{-0.2} N, \SI{0.2} N]$ and the range of frequency is $[\SI{0} \hertz, \SI{0.3} \hertz]$. This choice of ranges for the random \textit{Sinusoidal} exploration wrench is motivated to reflect motor babbling behavior a robotic end-effector will perform while manipulating an articulated object. The exploration wrench is applied for a duration of $\SI{5} \second$ and when the object is moving, we record the simulation time, links 3D pose values and the wrench values acting on the terminal links. Currently, our articulation model estimation algorithm \ref{alg:joint estimation} is offline and the recorded \textit{trial data} is passed as input.

In our modeling, the net wrench acting on the articulated object is given by,
\begin{equation}
  W= - X_{left}^* \ \mathrm{f}_{left} - X_{right}^* \ \mathrm{f}_{right} +  X_{com_1}m_1 \ \text{g} + X_{com_2}m_2 \ \text{g}
\label{TotalWrench}
\end{equation}

As our simplified model contains only one joint we have two sets, that represent the topology of the articulated object i.e.
\begin{subequations}
    \begin{equation}
        \Delta_{rev} = \{\{1,R\}\} \notag
    \end{equation}
    \begin{equation}
        \Delta_{pri} = \{\{1,P\}\} \notag
    \end{equation}
\end{subequations}

For each trial, we compute the following two model hypothesis error values for each of the objects,
\begin{itemize}
  \item \textit{Revolute Model Hypothesis Error}, which is the value that corresponds to the mismatch between the actual revolute joint motion and the revolute model hypothesis, given by,
  \begin{equation}
    R_{hyp}= \sum\limits_{data}||W - {^{\Delta_{rev}}{\dot{h}}}||
    \label{rhyp}
  \end{equation}
  
  \item \textit{Prismatic Model Hypothesis Error}, which is the value that corresponds to the mismatch between the actual prismatic joint motion and the prismatic articulation model hypothesis, given by,
  \begin{equation}
    P_{hyp}= \sum\limits_{data}||W - {^{\Delta_{pri}}{\dot{h}}}||
    \label{phyp}
  \end{equation}
\end{itemize}

\begin{figure}[h]
    \centering
    \includegraphics[scale=0.7]{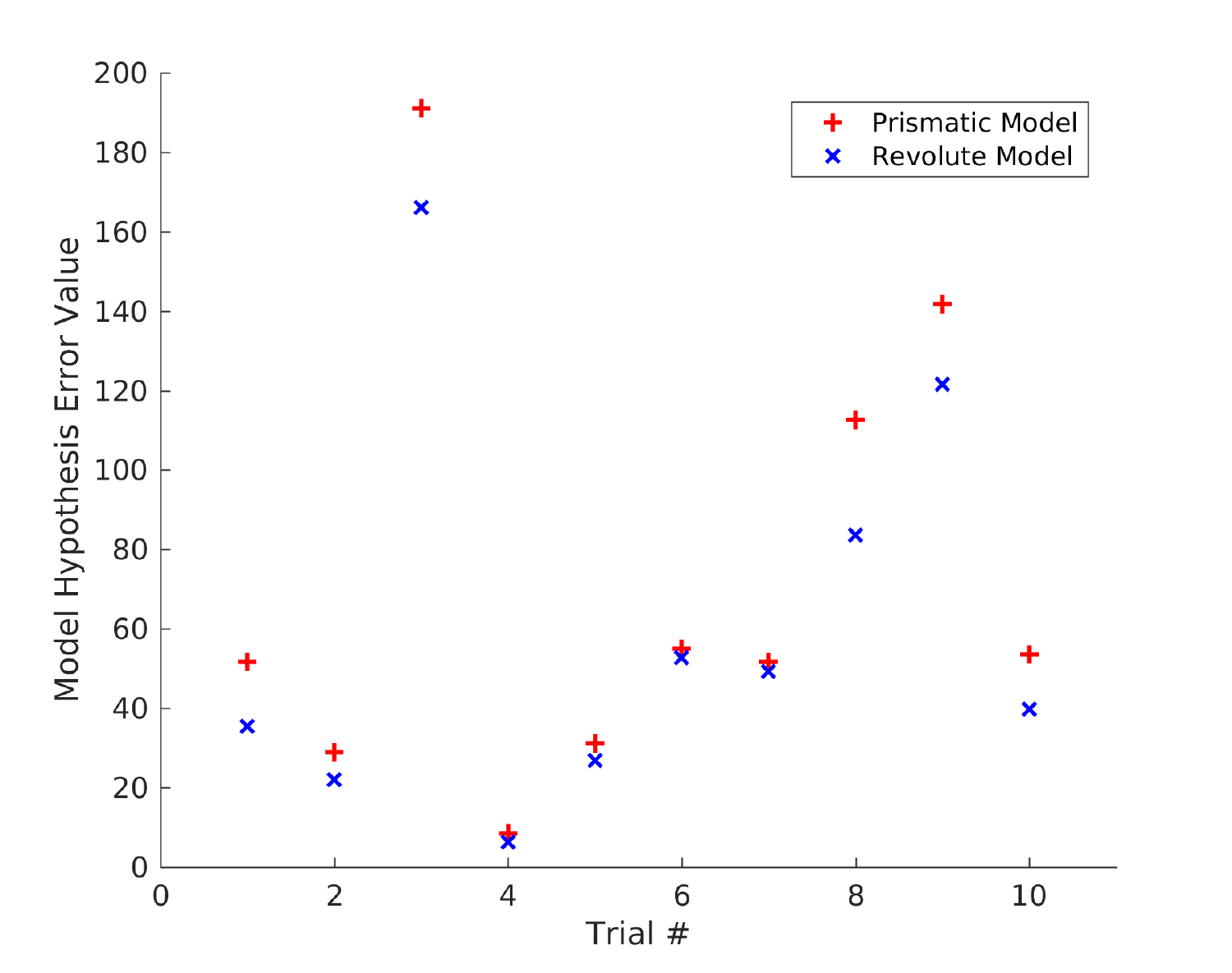}
    \caption{Model hypothesis error values of revolute joint}
    \label{rhypdiff-exp}
\end{figure}

\begin{figure}[h]
    \centering
    \includegraphics[scale=0.7]{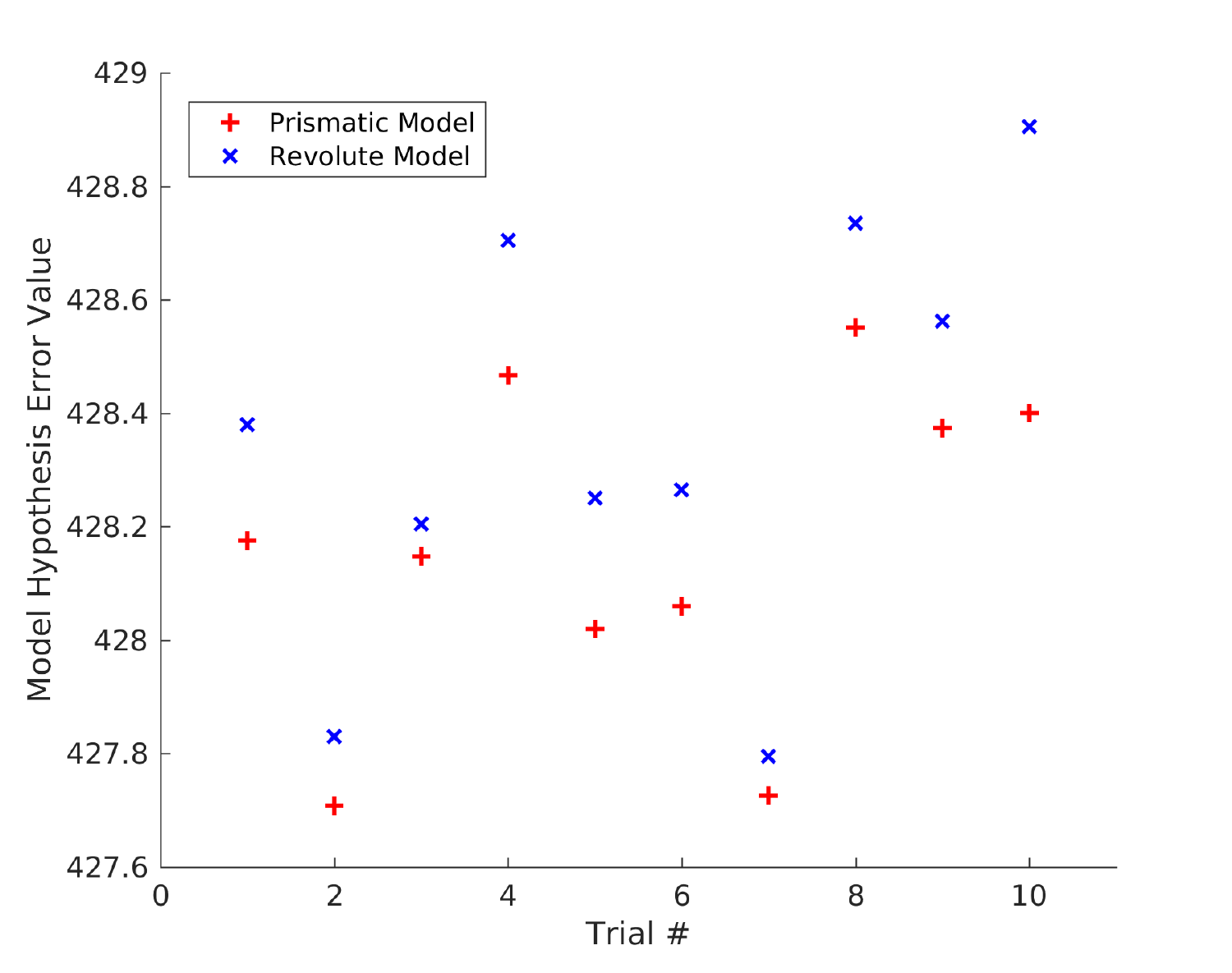}
    \caption{Model hypothesis error values of prismatic joint}
    \label{phypdiff-exp}
\end{figure}

The true topology of the articulated object ${\Delta}^*$ corresponds to the smallest model hypothesis error value. We ran several trials with random exploration wrench on the two articulated objects. As our algorithm depends on the kinematic evolution data, any wrench applied in the constrained direction results in $zero$ hypothesis error values. The model hypothesis error values of $10$ trials, in which the exploration wrench acted in the motion direction of the joint, is shown in Fig. \ref{rhypdiff-exp} and Fig. \ref{phypdiff-exp}. In the case of manipulating the articulated object with a revolute joint, the value of revolute model hypothesis error is less than the value of prismatic model hypothesis error as shown in Fig. \ref{rhypdiff-exp}. Similarly, in the case of manipulating the articulated object with a prismatic joint, the value of prismatic model hypothesis error is less than the value of revolute model hypothesis error as shown in Fig. \ref{phypdiff-exp}.

%% file: sections/conclusions.tex
\section{CONCLUSIONS}
In this paper we presented a general algorithm to estimate the topology of a multiple degrees of freedom articulated object. To demonstrate our algorithm with a simple case study, we made certain assumptions about the availability of rigid body pose in simulation and inertial parameters. A fixed time-step simulator like \textit{ode} ($1~ms$) adds considerable numerical errors for a stiff mechanical system used in our experiments. In such a noisy environment, our work is a proof of concept proving that we can estimate the articulation models using the kinematic evolution and interaction wrench information available during manipulation.

%% file: sections/acknowledgements.tex
\section{ACKNOWLEDGEMENTS}
This work is supported by \href{http://itn-pace.eu/}{PACE} project which has received funding from the European Union\textquotesingle s Horizon       2020 research and innovation programme under the Marie Sklodwska-Curie grant agreement No 642961.

%% file: sections/algorithm.tex
\begin{algorithm}[ht!]
    \caption{ Topology Estimation}
    \label{alg:joint estimation}
    \begin{algorithmic}[1]
        \REQUIRE: $t, \textit{P}_1, \textit{P}_2, \textit{P}_i,\textit{P}_{i+1},....,\textit{P}_n, \mathrm{f}_{left},\mathrm{f}_{right}$ 
  
        \WHILE{\texttildelow EOF}
        
           \STATE $dt$ $\leftarrow$ $diff(t)$
           
            \FOR{$i = 1:n$}
                
                \STATE $T_{i-1}$ $\leftarrow$ $H(P_{i-1})$
                \STATE $T_{i}$ $\leftarrow$ $H(P_{i})$
                \STATE $^iT_{i}$ $\leftarrow$ $T_{i-1}^{-1}$ $T_{i}$
                    
                \STATE $^{i-1}{R}_{i}$ $\leftarrow$ $rot(^{i-1}T_{i})$
                \STATE $axisAngleVector$ $\leftarrow$ $vrrotmat2vec(^{i-1}R_{i})$
                \STATE $^{rev}q_{i-1}$ $\leftarrow$ $axisAngleVector(4)$
                \STATE $^{rev}\mathrm{S}_{i-1}$ $\leftarrow$ $\begin{bmatrix}
                                                  0,0,0,axisAngleVector(1:3)
                                                \end{bmatrix}^T$
                \STATE $^{rev}\dot{q}_{i-1}$ $\leftarrow$ $\frac{diff(^{rev}q_{i-1})}{dt}$   
                \STATE $^{rev}\mathrm{v}_{J_{i-1}}$ $\leftarrow$ $X(T_{i-1})$ $^{rev}\mathrm{S}_{i-1} \ ^{rev}\dot{q}_{i-1}$
                    
                \STATE $^{i-1}p_{i}$ $\leftarrow$ $linear(^{i-1}T_{i})$
                \STATE $^{pri}q_{i-1}$ $\leftarrow$ $norm(^{i-1}p_{i})$
                \STATE $^{pri}\mathrm{S}_{i-1}$ $\leftarrow$ $\begin{bmatrix}
                                                   \frac{^{i-1}p_{i}}{^{pri}q_{i-1}},0,0,0
                                                \end{bmatrix}^T$
                \STATE $^{pri}\dot{q}_{i-1}$ $\leftarrow$ $\frac{diff(^{pri}q_{i-1})}{dt}$ 
                \STATE $^{pri}\mathrm{v}_{J_{i-1}}$ $\leftarrow$ $X(T_{i-1})$ $^{pri}\mathrm{S}_{i-1} \ ^{pri}\dot{q}_{i-1}$   
                
            \ENDFOR
            
            \STATE $W$ $\leftarrow$ $ {X}^*_{left}$ $\text{f}_{left} + {X}^*_{right}$ $\text{f}_{right} + \sum\limits_{i=1}^n {X_{{com}_{i}}^*} m_{i} \ \text{g}$
            
            \FOR{$j = 1:2^n$}
            
                \FOR{$i = 1:n$}
                    
                    \IF{$\Delta(i) = pri$}
                    
                        \IF{$\Delta(i-1) = pri$}
                            \STATE $^{pri}\mathrm{v}_{i+1} = \ ^{pri}\mathrm{v}_i + \  ^{pri}\mathrm{v}_{J_i}$
                        \ELSE
                            \STATE $^{pri}\mathrm{v}_{i+1} = \ ^{rev}\mathrm{v}_i + \  ^{pri}\mathrm{v}_{J_i}$
                        \ENDIF
                        
                        \STATE $^{pri}h_{i+1}$ $\leftarrow$ $[{X}_{i}^* \ M_{i} \ ^{i}{X}] \ ^{pri}\mathrm{v}_{i}$
                        \STATE $^{\Delta_j}h$ $\leftarrow$ $^{\Delta_j}h$ + $^{pri}h_{i+1}$ 
                    \ELSE{}
                        
                        \IF{$\Delta(i-1) = pri$}
                            \STATE $^{rev}\mathrm{v}_{i+1} = \ ^{pri}\mathrm{v}_i + ^{rev}\mathrm{v}_{J_i}$
                        \ELSE
                            \STATE $^{rev}\mathrm{v}_{i+1} = \ ^{rev}\mathrm{v}_i + ^{rev}\mathrm{v}_{J_i}$
                        \ENDIF
                        
                        \STATE $^{rev}h_{i+1}$ $\leftarrow$ $[{X}_{i}^* \ M_{i} \ ^{i}{X}] \ ^{rev}\mathrm{v}_{i}$
                        \STATE $^{\Delta_j}h$ $\leftarrow$ $^{\Delta_j}h$ + $^{rev}h_{i+1}$
                    \ENDIF
                        
                \ENDFOR
                
                \STATE $^{\Delta_j}\dot{h}$ $\leftarrow$ $\frac{diff(^{\Delta_j}h)}{dt}$
                
                \STATE $\Delta_j$ $\leftarrow$ $W$ $-$ $^{\Delta_j}\dot{h}$
                
            \ENDFOR
            
        \ENDWHILE

    \STATE ${\Delta}^*= \argmin\limits_{\Delta_j}\sum\limits^{2^n}_{j=1} ||W - {^{\Delta_j}{\dot{h}}}||$

    \end{algorithmic}
\end{algorithm}

%% file: main.bbl
\begin{thebibliography}{10}

\bibitem{nagatani1995experiment}
Keiji Nagatani and SI~Yuta.
\newblock An experiment on opening-door-behavior by an autonomous mobile robot
  with a manipulator.
\newblock In {\em Intelligent Robots and Systems 95.'Human Robot Interaction
  and Cooperative Robots', Proceedings. 1995 IEEE/RSJ International Conference
  on}, volume~2, pages 45--50. IEEE, 1995.

\bibitem{niemeyer1997simple}
G{\"u}nter Niemeyer and J-JE Slotine.
\newblock A simple strategy for opening an unknown door.
\newblock In {\em Robotics and Automation, 1997. Proceedings., 1997 IEEE
  International Conference on}, volume~2, pages 1448--1453. IEEE, 1997.

\bibitem{jain2009pulling}
Advait Jain and Charles~C Kemp.
\newblock Pulling open novel doors and drawers with equilibrium point control.
\newblock In {\em Humanoid Robots, 2009. Humanoids 2009. 9th IEEE-RAS
  International Conference on}, pages 498--505. IEEE, 2009.

\bibitem{karayiannidis2016adaptive}
Yiannis Karayiannidis, Christian Smith, Francisco Eli~Vina Barrientos, Petter
  {\"O}gren, and Danica Kragic.
\newblock An adaptive control approach for opening doors and drawers under
  uncertainties.
\newblock {\em IEEE Transactions on Robotics}, 32(1):161--175, 2016.

\bibitem{katz2007interactive}
Dov Katz and Oliver Brock.
\newblock Interactive perception: Closing the gap between action and
  perception.
\newblock In {\em ICRA 2007 Workshop: From features to actions-Unifying
  perspectives in computational and robot vision}, 2007.

\bibitem{katz2008manipulating}
Dov Katz and Oliver Brock.
\newblock Manipulating articulated objects with interactive perception.
\newblock In {\em Robotics and Automation, 2008. ICRA 2008. IEEE International
  Conference on}, pages 272--277. IEEE, 2008.

\bibitem{brock2009learning}
Dov Katz Yuri Pyuro~Oliver Brock.
\newblock Learning to manipulate articulated objects in unstructured
  environments using a grounded relational representation.
\newblock {\em Robotics: Science and Systems IV}, page 254, 2009.

\bibitem{sturm2009learning}
J{\"u}rgen Sturm, Vijay Pradeep, Cyrill Stachniss, Christian Plagemann, Kurt
  Konolige, and Wolfram Burgard.
\newblock Learning kinematic models for articulated objects.
\newblock In {\em IJCAI}, pages 1851--1856, 2009.

\bibitem{sturm20103d}
J{\"u}rgen Sturm, Kurt Konolige, Cyrill Stachniss, and Wolfram Burgard.
\newblock 3d pose estimation, tracking and model learning of articulated
  objects from dense depth video using projected texture stereo.
\newblock In {\em RGB-D: Advanced Reasoning with Depth Cameras Workshop, RSS},
  2010.

\bibitem{sturm2011probabilistic}
J{\"u}rgen Sturm, Cyrill Stachniss, and Wolfram Burgard.
\newblock A probabilistic framework for learning kinematic models of
  articulated objects.
\newblock {\em Journal of Artificial Intelligence Research}, 41:477--526, 2011.

\bibitem{hausman2015active}
Karol Hausman, Scott Niekum, Sarah Osentoski, and Gaurav~S Sukhatme.
\newblock Active articulation model estimation through interactive perception.
\newblock In {\em Robotics and Automation (ICRA), 2015 IEEE International
  Conference on}, pages 3305--3312. IEEE, 2015.

\bibitem{otte2014entropy}
Stefan Otte, Johannes Kulick, Marc Toussaint, and Oliver Brock.
\newblock Entropy-based strategies for physical exploration of the
  environment's degrees of freedom.
\newblock In {\em Intelligent Robots and Systems (IROS 2014), 2014 IEEE/RSJ
  International Conference on}, pages 615--622. IEEE, 2014.

\bibitem{martin2014online}
Roberto~Martin Martin and Oliver Brock.
\newblock Online interactive perception of articulated objects with multi-level
  recursive estimation based on task-specific priors.
\newblock In {\em Intelligent Robots and Systems (IROS 2014), 2014 IEEE/RSJ
  International Conference on}, pages 2494--2501. IEEE, 2014.

\bibitem{martin2017building}
Roberto Mart{\i}n-Mart{\i}n and Oliver Brock.
\newblock Building kinematic and dynamic models of articulated objects with
  multi-modal interactive perception.
\newblock In {\em AAAI Symposium on Interactive Multi-Sensory Object Perception
  for Embodied Agents, AAAI, Ed}, 2017.

\bibitem{featherstone2014rigid}
Roy Featherstone.
\newblock {\em Rigid body dynamics algorithms}.
\newblock Springer, 2014.

\bibitem{featherstone2014rigidCh3}
Roy Featherstone.
\newblock {\em Rigid body dynamics algorithms}, chapter~3, pages 49--50.
\newblock Springer, 2014.

\bibitem{featherstone2014rigidCh2}
Roy Featherstone.
\newblock {\em Rigid body dynamics algorithms}, chapter~2, pages 35--36.
\newblock Springer, 2014.

\bibitem{metta2010icub}
Giorgio Metta, Lorenzo Natale, Francesco Nori, Giulio Sandini, David Vernon,
  Luciano Fadiga, Claes Von~Hofsten, Kerstin Rosander, Manuel Lopes, Jos{\'e}
  Santos-Victor, et~al.
\newblock The icub humanoid robot: An open-systems platform for research in
  cognitive development.
\newblock {\em Neural Networks}, 23(8-9):1125--1134, 2010.

\bibitem{Nataleeaaq1026}
Lorenzo Natale, Chiara Bartolozzi, Daniele Pucci, Agnieszka Wykowska, and
  Giorgio Metta.
\newblock icub: The not-yet-finished story of building a robot child.
\newblock {\em Science Robotics}, 2(13), 2017.

\bibitem{nori2015icub}
Francesco Nori, Silvio Traversaro, Jorhabib Eljaik, Francesco Romano, Andrea
  Del~Prete, and Daniele Pucci.
\newblock icub whole-body control through force regulation on rigid
  non-coplanar contacts.
\newblock {\em Frontiers in Robotics and AI}, 2:6, 2015.

\bibitem{hoffman2014yarp}
Enrico~Mingo Hoffman, Silvio Traversaro, Alessio Rocchi, Mirko Ferrati,
  Alessandro Settimi, Francesco Romano, Lorenzo Natale, Antonio Bicchi,
  Francesco Nori, and Nikos~G Tsagarakis.
\newblock Yarp based plugins for gazebo simulator.
\newblock In {\em International Workshop on Modelling and Simulation for
  Autonomous Systems}, pages 333--346. Springer, 2014.

\end{thebibliography}
